# A SINGLE-SHOT OBJECT DETECTOR WITH FEATURE AGGREGATION AND ENHANCEMENT

*Weiqiang Li, Guizhong Liu*

Institute of Multimedia Processing and Communication, Ministry of Education Key Lab for Intelligent Networks and Network Security, Xi'an Jiaotong University, Xi'an, China 710049
liugz@xjtu.edu.cn

## ABSTRACT

For many real applications, it's equally important to detect objects accurately and quickly. In this paper, we propose an accurate and efficient single shot object detector with feature aggregation and enhancement (FAENet). Our motivation is to enhance and exploit the shallow and deep feature maps of the whole network simultaneously. To achieve it we introduce a pair of novel feature aggregation modules and two feature enhancement blocks, and integrate them into the original structure of SSD. Extensive experiments on both the PASCAL VOC and MS COCO datasets demonstrate that the proposed method achieves much higher accuracy than SSD. In addition, our method performs better than the state-of-the-art one-stage detector RefineDet on small objects and can run at a faster speed.

## 1. INTRODUCTION

In recent years, object detection has been promoted significantly with the rapid development of deep neural networks. Currently, object detectors can be divided into two series: the region based two-stage frameworks such as [1, 2, 3, 4] and the region free one-stage frameworks such as [5, 6, 7]. Generally speaking, the two-stage detectors hold higher accuracy on challenging datasets such as those in [8, 9] while the one-stage detectors have higher inference speeds.

For their higher implementation speed, the one-stage object detectors have received much attention recently. Authors of [6] proposed an object detector named SSD (Single Shot MultiBox Detector), which has become the baseline of most newly proposed one-stage object detectors. SSD first generates some low-level feature maps by a backbone [10], then adds several consecutive convolutional layers to extract high-level feature maps with more semantic

This work is supported by Joint Foundation of Ministry of Education of China (No. 6141A020223).

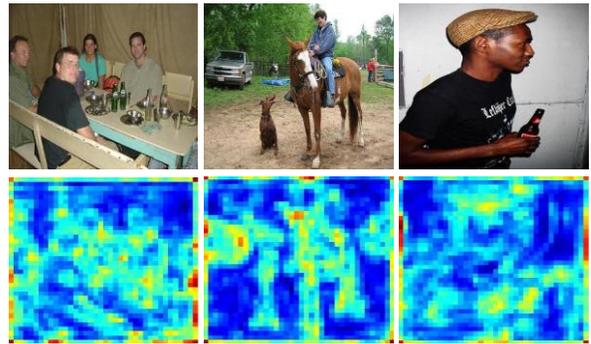

**Fig. 1**. Illustrations on some images and their corresponding Conv4_3 feature maps in SSD [6].

information. It uses lower level features to detect smaller objects and larger objects are detected by higher level feature maps. However, the lower layers extract less semantic information so that their features might not have enough capability to detect small objects.

In this paper, we first point out the following three problems existing in SSD: (1) As shown in Fig. 1, the Conv4_3 is a shallow layer whose features lack semantics to discriminate smaller objects like bottles directly; (2) There are small objects with large aspect ratios in several classes such as boat and bottle; (3) Although SSD predicts objects by multi-scale feature maps, they are operated separately so that information in different layers is not made full use of.

In order to address the problems mentioned above, we design a new one-stage object detection architecture named FAENet. It improves the detection accuracy and maintains a faster inference speed, which is beneficial for real applications. This is achieved by introducing a pair of novel feature aggregation modules and two feature enhancement blocks to the original SSD. Extensive experiments on the datasets PASCAL VOC and MS COCO show that our method performs much better than SSD and is superior in detecting small objects to the state-of-the-art one-stage object detector in [11]. For instance, for the input image size of 300×300, FAENet gains 80.1% mAP and runs at 63.5 FPS on VOC 2007. In addition, FAENet achieves 16.0% mAP for small objects on MS COCO.

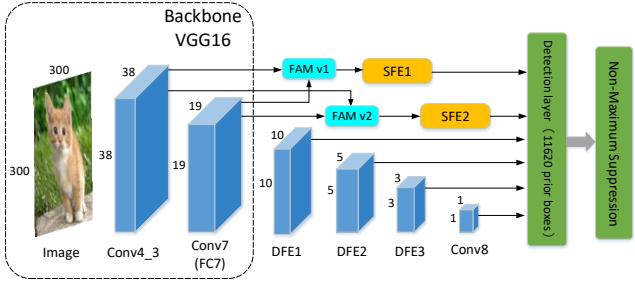

**Fig. 2**. The overall network architecture of FAENet with the input size 300×300.

## 2. RELATED WORK

There are two classes of object detectors based on deep neural networks: the two-stage detectors and the one-stage detectors. A two-stage detector first generates a set of object region proposals by some methods such as those in [12, 13, 2], and then the region proposals are further classified and regressed in the second stage.

A one-stage detector eliminates the proposal generation step and carries out classification and bounding box regression in densely sampled boxes directly. Among the one-stage detectors, semantic feature enhancement and multi-level feature fusion are two common means to improve small objects detection. Authors in [7] combined SSD with a stronger backbone network [14], and constructed an hourglass network by introducing several deconvolutional layers to enhance low-level semantic information. Literature [15] used the last stage of [16] to construct a feature pyramid by pooling and scale-transfer layers. Literature [11] used transfer connection blocks to build a feature pyramid by fusing the deep layers and shallow layers in a top-down manner. Reference [17] enriched the semantics of low-level features by adding a semantic segmentation branch.

Our work also adopts the above mentioned two effective means for improving object detection. However, we use two different feature enhancement blocks in shallow and deep layers respectively. In addition, we design a pair of novel feature aggregation modules to merge the features of Conv4_3 and FC7. Without the feature pyramid in a top-down manner, our method can also detect small objects well while maintaining a high computational efficiency.

## 3. PROPOSED METHOD

As illustrated in Fig. 2, the proposed FAENet extends the detection framework of SSD. We use two well-designed Feature Aggregation Modules (FAM) to combine feature maps of Conv4_3 and FC7, which have different semantic information, then adds Shallow Feature Enhancement (SFE) blocks to further strengthen the semantics of these two layers respectively. As for the deeper layers, we replace the additional convolutional layers in the original SSD with three Deep Feature Enhancement (DFE) blocks. In the following

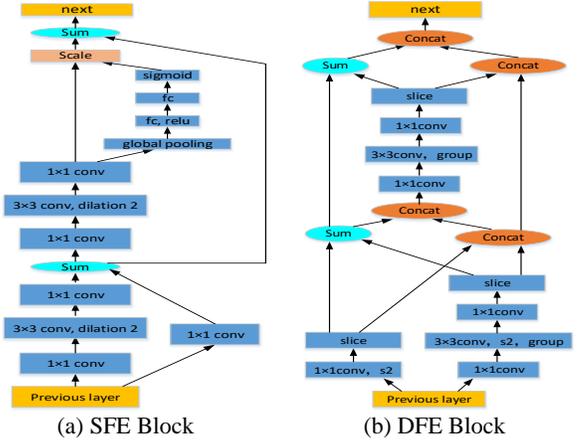

(a) SFE Block  (b) DFE Block

**Fig. 3**. The structure of two Feature Enhancement Blocks used in our FAENet, in which each conv box denotes a BN+ReLu+Conv processing. (a) SFE Block. (b) DFE Block.

sub-sections, we will explain these core components in detail.

### 3.1 Feature Enhancement Block

In the original SSD, the shallower layers like Conv4_3 do not have rich semantic information so that there are problems with small objects detection. Motivated by [14, 18, 19, 20], we design Shallow Feature Enhancement (SFE) Blocks for Conv4_3 and FC7. Specifically, as shown in Fig. 3.a, we use two consecutive Residual Units to deepen the shallow layers and learn more nonlinear representations. In each bottleneck, we use a 3×3 conv with dilation rate of 2 which can broaden the receptive field. Larger receptive field means more contextual information can be learned, which is helpful for detecting objects, according to [21]. In the second Residual Unit, the SE Block in [20] is added to utilize its function of recalibrating different channels, making enhanced features more adaptive.

For deep layers, SSD only adds a few extra convolutional layers, which we think are not enough representative, and the deeper feature maps may lose numerous details about the input image. Therefore, we use the basic unit in [22] as our Deep Feature Enhancement (DFE) Block, which combines the advantages of both [14] and [16]. The detailed structure is shown in Fig. 3.b. It not only deepens the whole network, but also has an implicit feature fusion from the low layers to the high layers, enabling higher features more informative. Ablation Study in Section 4 verifies the effectiveness of these two Feature Enhancement Blocks.

### 3.2. Feature Aggregation Module

In order to fuse the feature maps of Conv4_3 and FC7, we design a pair of Feature Aggregation Modules (FAM) and use them in these two layers respectively, as can be seen in Fig. 4.a and Fig. 4.b.

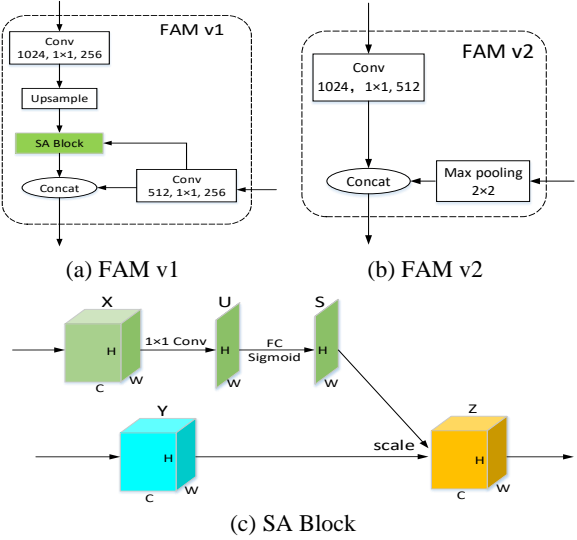

**Fig. 4**. The structure of our Feature Aggregation Modules (FAM), in which each Conv box denotes a Conv+BN+ReLu processing. (a) FAM v1. (b) FAM v2. (c) SA Block.

Unlike element-wise sum used in some top-down architectures, we use concatenation to aggregate feature maps from different layers. The feature maps in FC7 usually have more semantics than those in Conv4_3 and also have a relatively reasonable resolution. According to this, we propose a Spatial Attention (SA) Block to highlight objective regions, which is favorable for detecting small objects.

As shown in Fig. 4.c, given two input features with the same number of channels: $X \in \mathbb{R}^{C \times H \times W}$ and $Y \in \mathbb{R}^{C \times H \times W}$, a 1×1 Conv is first used to aggregate multiple channels into one and produce $U \in \mathbb{R}^{H \times W}$ by:

$$U = F_{Conv}(X) \quad (1)$$

where $F_{Conv}(\cdot)$ represents a 1×1Conv+BN+ReLu operation. Then we use a fully connected layer followed by an activation function to generate weight activation values at each spatial location:

$$S = \sigma(WU) \in \mathbb{R}^{H \times W} \quad (2)$$

where $\sigma$ refers to the sigmoid function. Finally, the output Z is obtained by reweighting the input Y:

$$Z_{ihw} = Y_{ihw} \cdot S_{hw} \quad (3)$$

where $Z \in \mathbb{R}^{C \times H \times W}$. Through this series of operations, the objective regions in shallow layers get effectively enhanced.

The whole Feature Aggregation Modules use features of Conv4_3 and FC7 to enhance and refine each other, which is very helpful for detecting small and medium objects.

In general, FAENet inherits the framework of SSD and adds to it a few independent and simple modules, without using the complicated top-down structure.

## 4. EXPERIMENTS

We conduct experiments on two datasets: PASCAL VOC 2007 and MS COCO, which include 20 and 80 categories of images respectively. We only use VGG-16 as backbone network for all the experiments. The evaluation metric for both datasets is the mean Average Precision (mAP). We conduct all the experiments with an NVIDIA GeForce GTX 1080Ti GPU and our code is based on PyTorch.

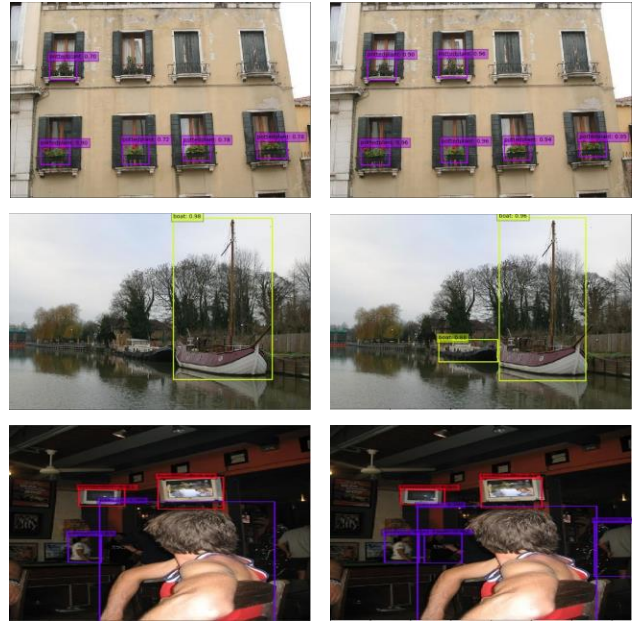

**Fig. 5**. Examples of detection results on PASCAL VOC 2007. Left: SSD300. Right: FAENet300.

### 4.1. PASCAL VOC 2007

In this experiment, we train our model on the union of VOC 2007 and VOC 2012 *trainval* sets, then test on VOC 2007 *test* set. We use two different input sizes: 300×300 and 512×512. For the input size of 300, we set the batch size to 32 and use a warm-up strategy to train for the first 5 epochs. We initialize the learning rate at $4 \times 10^{-3}$, divided by 10 at the 150-th and 200-th epochs respectively and stop training at 250-th epochs. For a larger input size of 512, we set the batch size to 12 due to memory limitations of GPU, and the other super-parameters are the same as those of the size 300.

As shown in Table 1, we compare FAENet with several other state-of-the-art one-stage detectors. For the input size of 300, FAENet achieves 80.1% mAP, exceeding SSD by 2.6 points and outperforms all the other detectors except RFBNet [23], whose mAP is 80.5%. Our method has the highest accuracy in some categories including bike, chair, cow, table and sofa. It implies that our method is superior in some scenes. As shown in Table 3, for a larger input size, FAENet still exceeds SSD by 1.3 points and also performs better than STDN which uses a deeper backbone. Moreover, our method possesses a very fast inference speed, it can run at 63.5 FPS when the input image size is 300×300, which is meaningful for applications requiring real-time processing.

In summary, our method is only slower than RFBNet in inference speed and has a higher detection accuracy than SSD and STDN. When the input image size is small, FAENet outperforms RefineDet and is only slightly inferior to RFBNet. Fig. 5 shows some examples of detection results.

**Table 1.** PASCAL VOC 2007 *test* detection results.

| Method | mAP | aero | bike | bird | boat | bottle | bus | car | cat | chair | cow | table | dog | horse | mbike | person | plant | sheep | sofa | train | tv |
|---|---|---|---|---|---|---|---|---|---|---|---|---|---|---|---|---|---|---|---|---|---|
| SSD300 [6] | 77.5 | 79.5 | 83.9 | 76.0 | 69.6 | 50.5 | 87.0 | 85.7 | 88.1 | 60.3 | 81.5 | 77.0 | 86.1 | 87.5 | 83.9 | 79.4 | 52.3 | 77.9 | 79.5 | 87.6 | 76.8 |
| STDN300 [15] | 78.1 | 81.1 | 86.9 | 76.4 | 69.2 | 52.4 | 87.7 | 84.2 | 88.3 | 60.2 | 81.3 | 77.6 | **86.6** | 88.9 | **87.8** | 76.8 | 51.8 | 78.4 | 81.3 | 87.5 | 77.8 |
| RefineDet320 [11] | 80.0 | 83.9 | 85.4 | **81.4** | 75.5 | 60.2 | 86.4 | **88.1** | **89.1** | 62.7 | 83.9 | 77.0 | 85.4 | 87.1 | 86.7 | **82.6** | 55.3 | **82.7** | 78.5 | 88.1 | 79.4 |
| RFBNet300 [23] | **80.5** | **85.0** | 86.1 | 77.7 | **75.7** | **60.6** | **88.9** | 87.6 | 86.8 | 64.2 | 85.3 | 77.9 | 86.1 | **89.0** | 87.1 | 82.2 | **58.7** | 81.5 | 81.1 | **88.3** | **81.5** |
| FAENet300 (ours) | 80.1 | 82.8 | **87.3** | 76.5 | 74.7 | 58.7 | 86.5 | 87.5 | 88.2 | **65.7** | **85.9** | **79.2** | 85.4 | 88.4 | 86.9 | 81.4 | 57.7 | 80.4 | **82.2** | 86.8 | 79.6 |

**Table 2.** Detection results on MS COCO *test-dev* 2015 set.

| Method | Data | Backbone | AP | $AP_{50}$ | $AP_{75}$ | $AP_S$ | $AP_M$ | $AP_L$ |
|---|---|---|---|---|---|---|---|---|
| SSD300 [6] | trainval35k | VGG-16 | 25.1 | 43.1 | 25.8 | 6.6 | 25.9 | 41.4 |
| STDN300 [15] | trainval35k | DenseNet-169 | 28.0 | 45.6 | 29.4 | 7.9 | 29.7 | 45.1 |
| RefineDet320 [11] | trainval35k | VGG-16 | 29.4 | 49.2 | 31.3 | 10.0 | 32.0 | 44.4 |
| RFBNet300 [23] | trainval35k | VGG-16 | 30.3 | 49.3 | 31.8 | 11.8 | 31.9 | 45.9 |
| FAENet300 (ours) | trainval35k | VGG-16 | 28.3 | 47.9 | 29.7 | 10.5 | 30.9 | 41.9 |
| SSD512 [6] | trainval35k | VGG-16 | 28.8 | 48.5 | 30.3 | 10.9 | 31.8 | 43.5 |
| STDN513 [15] | trainval35k | DenseNet-169 | 31.8 | 51.0 | 33.6 | 14.4 | 36.1 | 43.4 |
| RefineDet512 [11] | trainval35k | VGG-16 | 33.0 | 54.5 | 35.5 | 16.3 | 36.3 | 44.3 |
| RFBNet512 [23] | trainval35k | VGG-16 | 33.8 | 54.2 | 35.9 | 16.2 | 37.1 | 47.4 |
| FAENet512 (ours) | trainval35k | VGG-16 | 31.8 | 51.2 | 33.5 | 16.0 | 35.8 | 42.7 |

**Table 3.** Comparison on PASCAL VOC 2007 *test* set.

| Method | Backbone | mAP | FPS |
|---|---|---|---|
| SSD300 [6] | VGG-16 | 77.5 | 46 |
| STDN300 [15] | DenseNet-169 | 78.1 | 41.5 |
| RefineDet320 [11] | VGG-16 | 80.0 | 40.3 |
| RFBNet300 [23] | VGG-16 | 80.5 | 83 |
| FAENet300 (ours) | VGG-16 | 80.1 | 63.5 |
| SSD512 [6] | VGG-16 | 79.8 | 19 |
| STDN513 [15] | DenseNet-169 | 80.9 | 28.6 |
| RefineDet512 [11] | VGG-16 | 81.8 | 24.1 |
| RFBNet512 [23] | VGG-16 | 82.2 | 38 |
| FAENet512 (ours) | VGG-16 | 81.1 | 32.8 |

**Table 4.** Ablation study on PASCAL VOC 2007.

| Component | SSD | | | FAE |
|---|---|---|---|---|
| + SFE | | ✓ | ✓ | ✓ |
| + DFE | | | ✓ | ✓ |
| + FAM | | | | ✓ |
| mAP | 77.5 | 79.0 | 79.5 | **80.1** |

### 4.2. MS COCO

We experiment on the more challenging MS COCO dataset to further validate our method. We use the *trainval35k* set for training and report the results from *test-dev* evaluation server. The batch size is set to 32 for the input size of 300 and 12 for the input size of 512 respectively. The learning rate is initialed at $2 \times 10^{-3}$ for both input sizes. We still use the warm-up strategy to train for the first 5 epochs, then divided by 10 at the 80-th and 110-th epochs respectively and stop training at the 140-th epochs.

As shown in Table 2, FAENet gains 28.3% mAP when the input size is 300×300, exceeding SSD by 3.2 points. We can also see that FAENet achieves 10.5% mAP on small objects, which performs better than SSD, STDN and RefineDet. For a larger input size of 512, our method gains 31.8% mAP, exceeding SSD by 3 points and equal to that of STDN which uses a deeper and stronger backbone. For small objects, FAENet can achieve 16.0% mAP, which is very close to the state-of-the-art one-stage detectors RFBNet and RefineDet.

### 4.3. Ablation Study

We conduct ablation study on PASCAL VOC 2007 to further verify the effectiveness of different components in FAENet. Table 4 shows the results of our ablation study. We first only use the Feature Enhancement Block (SFE Block) in shallow layers and the result is 79.0%. Then we introduce DFE Block to deep layers and find that mAP rises by 0.5 points. Finally, after adding two FAMs, we build the whole FAENet, whose mAP is improved from 79.5% to 80.1%. These ablation experiments prove that each element of FAENet helps to improve detection accuracy.

## 5. CONCLUSION AND FUTURE WORK

In this paper, we present an accurate and efficient one-stage object detector so called FAENet. A pair of FAMs are introduced to improve small objects detection by fusing features of different layers. In addition, we utilize the SFE Block and the DFE Block to enhance semantic information of shallow features and detail information of deep features respectively. The experimental results on PASCAL VOC and MS COCO demonstrate that our model excels both in accuracy and speed, particularly for small objects. In the future we expect to use deeper and stronger backbone networks, which may be more beneficial for detecting larger objects, and study their impact on detecting objects at different scales.

# 6. REFERENCES

[1] R. Girshick, J. Donahue, T. Darrell, and J. Malik. Rich feature hierarchies for accurate object detection and semantic segmentation. In *CVPR*, pages 580-587, 2014.

[2] S. Ren, K. He, R. Girshick, and J. Sun. Faster R-CNN: towards real-time object detection with region proposal networks. In *NIPS*, pages 91–99, 2015.

[3] J. Dai, Y. Li, K. He, and J. Sun. R-FCN: object detection via region-based fully convolutional networks. In *NIPS*, pages 379-387, 2016.

[4] T. Lin, P. Dollár, R. B. Girshick, K. He, B. Hariharan, and S. J. Belongie. Feature pyramid networks for object detection. In *CVPR*, pages 936-944, 2017.

[5] J. Redmon, S. K. Divvala, R. B. Girshick, and A. Farhadi. You only look once: Unified, real-time object detection. In *CVPR*, pages 779-788, 2016.

[6] W. Liu, D. Anguelov, D. Erhan, C. Szegedy, S. E. Reed, C.Fu, and A. C. Berg. SSD: single shot multibox detector. In *ECCV*, pages 21-37, 2016.

[7] C. Fu, W. Liu, A. Ranga, A. Tyagi, and A. C. Berg. DSSD: Deconvolutional single shot detector. *arXiv preprint arXiv:1701.06659*, 2017.

[8] M. Everingham, L. J. V. Gool, C. K. I. Williams, J. M. Winn, and A. Zisserman. The PASCAL Visual Object Classes (VOC) challenge. *IJCV*, 88(2):303–338, 2010.

[9] T. Lin, M. Maire, S. J. Belongie, J. Hays, P. Perona, D. Ramanan, P. Dollár, and C. L. Zitnick. Microsoft COCO: common objects in context. In *ECCV*, pages 740-755, 2014.

[10] K. Simonyan and A. Zisserman. Very deep convolutional networks for large-scale image recognition. *CoRR*, abs/1409.1556, 2014.

[11] S. Zhang, L. Wen, X. Bian, Z. Lei, and S. Z. Li. Single-shot refinement neural network for object detection. In *CVPR*, pages 4203-4212, 2018.

[12] J. R. R. Uijlings, K. E. A. van de Sande, T. Gevers, and A. W. M. Smeulders. Selective search for object recognition. *IJCV*, 104(2):154–171, 2013.

[13] C. L. Zitnick and P. Dollár. Edge boxes: Locating object proposals from edges. In *ECCV*, pages 391-405, 2014.

[14] K. He, X. Zhang, S. Ren, and J. Sun. Deep residual learning for image recognition. In *CVPR*, pages 770-778, 2016.

[15] P. Zhou, B. Ni, C. Geng, J. Hu, and Y. Xu. Scale-transferrable object detection. In *CVPR*, pages 528-537, 2018.

[16] G. Huang, Z. Liu, L. van der Maaten, and K. Q. Weinberger. Densely connected convolutional networks. In *CVPR*, pages 2261-2269, 2017.

[17] Z. Zhang, S. Qiao, C. Xie, W. Shen, B. Wang, and A. Yuille. Single-shot object detection with enriched semantics. In *CVPR*, pages 5813-5821, 2018.

[18] W. Luo, Y. Li, R. Urtasun, and R. Zemel. Understanding the effective receptive field in deep convolutional neural networks. In *NIPS*, pages 4898-4906, 2016.

[19] F. Yu and V. Koltun. Multi-scale context aggregation by dilated convolutions. *arXiv preprint arXiv:1511.07122*, 2015.

[20] J. Hu, L. Shen, and G. Sun. Squeeze-and-excitation networks. In *CVPR*, pages 7132-7141, 2018.

[21] A. Torralba. Contextual priming for object detection. *IJCV*, 53(2):169-191, 2003.

[22] Y. Chen, J. Li, H. Xiao, X. Jin, S. Yan, and J. Feng. Dual path networks. In *NIPS*, pages 4470-4478, 2017.

[23] S. Liu, D. Huang, and Y. Wang. Receptive field block net for accurate and fast object detection. In *ECCV*, pages 385-400, 2018.